\documentclass[conference]{STPINN}
\IEEEoverridecommandlockouts
\usepackage{cite}
\usepackage{amsmath,amssymb,amsfonts}
\usepackage{algorithmicx}
\usepackage{graphicx}
\usepackage{textcomp}
\usepackage{xcolor}
\usepackage{algorithm}
\usepackage{algpseudocode}
\usepackage{booktabs}
\usepackage{multirow}

\begin{document}

\title{ST-PINN: A Self-Training Physics-Informed Neural Network for Partial Differential Equations \thanks{This research work was supported in part by the National Key Research and Development Program of China (2021YFB0300101). The paper is accepted by International Joint Conference on Neural Networks (IJCNN2023).}}

\author{
    \IEEEauthorblockN{Junjun Yan$^1$, Xinhai Chen$^{1,*}$, Zhichao Wang$^1$, Enqiang Zhoui$^2$ and Jie Liu$^2$}
    \IEEEauthorblockA{$^1$ Science and Technology on Parallel and Distributed Processing Laboratory,\\ National University of Defense Technology, Changsha, 410073, China}
    \IEEEauthorblockA{$^2$ Laboratory of Digitizing Software for Frontier Equipment,\\ National University of Defense Technology, Changsha, 410073, China}
    \IEEEauthorblockA{$^*$ Corresponding author: chenxinhai16@nudt.edu.cn}
 }

\maketitle

\begin{abstract}
    Partial differential equations (PDEs) are an essential computational kernel in physics and engineering. With the advance of deep learning, physics-informed neural networks (PINNs), as a mesh-free method, have shown great potential for fast PDE solving in various applications. To address the issue of low accuracy and convergence problems of existing PINNs, we propose a self-training physics-informed neural network, ST-PINN. Specifically, ST-PINN introduces a pseudo label based self-learning algorithm during training. It employs governing equation as the pseudo-labeled evaluation index and selects the highest confidence examples from the sample points to attach the pseudo labels. To our best knowledge, we are the first to incorporate a self-training mechanism into physics-informed learning. We conduct experiments on five PDE problems in different fields and scenarios. The results demonstrate that the proposed method allows the network to learn more physical information and benefit convergence. The ST-PINN outperforms existing physics-informed neural network methods and improves the accuracy by a factor of 1.33x-2.54x. The code of ST-PINN is available at GitHub: https://github.com/junjun-yan/ST-PINN.
\end{abstract}


\section{Introduction}

Partial differential equations (PDEs) are crucial in physics and engineering, such as computational fluid dynamics, electromagnetic theory, and quantum mechanics \cite{ns1, HFM, nd1}. However, the numerical methods are sometimes computationally expensive in inverse problem solving, complex geometry domain solving, and high-dimension space solving \cite{phy-inf}. With the emergence of deep learning, some data-driven models have successfully been applied in the physics and engineering fields \cite{pinnrev, reveiw1, reveiw2}. Despite the ability of data-driven models to establish a function map between input and output data, their accuracy is closely related to the size and distribution of the data. Furthermore, the data-driven models overlook the prior physical knowledge of physics and engineering problems, which is a waste of information.

Many researchers have addressed the above issue by combining prior physics knowledge with data-driven models. This approach, known as physics-informed learning \cite{pinnrev, phy-inf}, has led to the development of physics-informed neural networks (PINNs), which embed PDEs into the loss function and convert numerical computation problems into optimization problems \cite{PINN, deepxde}. PINNs require only a small amount of (or even no) supervised data to train. Moreover, PINNs are mesh-free, which means they can randomly sample points in the domain as unlabeled training data without generating mesh. While theoretical analyses suggest that PINNs can converge to solutions in some situations, their accuracy is still inadequate in practical applications \cite{theory2, shin, mishra, theory1}.

Semi-Supervised Learning (SSL) is a machine learning method between data-driven supervised learning and unsupervised learning. It is suited for scenarios where only a small amount of labeled data is available but a large amount of unlabeled data \cite{semi1, semi2, semi4}. SSL methods make full use of the unlabeled data to improve the accuracy of the models. One of the most widely used SSL methods is self-training, which assigns pseudo labels to the most confident unlabeled samples and includes these pseudo labels in supervised training \cite{semi5, semi3}. However, the most challenging part of this process is selecting the unlabeled samples to assign pseudo labels. Although self-training has successfully been applied in many fields, there have been few studies in the context of physics-informed learning \cite{semi1, semi4}.

In this paper, we proposed ST-PINN, which combines self-training with physics-informed neural networks (PINNs) to leverage unlabeled data and physical information. Our work is motivated by two key factors. Firstly, in many practical applications, obtaining the full range of physics data in the training domain can be expensive. Thus, only a few observed data points are available, which aligns with the SSL scenario. Secondly, the residual loss of the physics equation is a natural criterion for selecting pseudo points since predicted values that conform to the equation are likely to be more accurate. 

Specifically, we begin by warming up the network with traditional training steps. After several iterations, the network predicts all the sample points and computes the residual loss of the physics equation. Then, ST-PINN selects some sample points with the minimum equation loss and assigns them pseudo labels. In the next iteration, the pseudo points will be incorporated as supervised data into training. At the end of every several iterations, we generate new pseudo points using the above processes. With an appropriate pseudo-generating strategy, the pseudo points can be considered an extension of supervised data, which benefits network convergence.

In general, our contributions can be summarized as follows:
\begin{itemize} 
    \item We propose ST-PINN, a pseudo label based self-training framework for training PINNs. To our best knowledge, this is the first research to combine self-training mechanisms and physics-informed learning. ST-PINN uses the physical equation to generate pseudo points and treats the pseudo points as a supervised data form, which makes full use of unlabeled data with physics information. 
    \item We design a strategy for pseudo-label generation and introduce three hyperparameters to control the quantity and quality of pseudo points. These three hyperparameters can stabilize the training process and trade-off efficiency and accuracy, which is essential in ST-PINN. 
    \item We conduct a series of experiments in five different PDEs. Compared with the original PINN, our model can improve the accuracy by about 1.33$\times$-2.54$\times$. The experimental results demonstrate that the self-training mechanism can benefit network convergence and improve prediction accuracy. 
\end{itemize}

The remainder paper is organized as follows: In section II, we introduce the related works and background. Next, we describe the details of our method and the pseudo points generation strategy in section III. In section IV, we show our experimental environment and evaluate our method in different PDEs. Finally, we give the conclusions in section V.

\section{Related Works}
Neural networks are one of the most well-known data-driven machine learning models widely used in various fields \cite{imagenet, cxh1}. Physics-informed neural networks (PINNs) are specific types of neural networks designed for solving physics problems. These networks not only learn from the distribution of supervised data but also aim to comply with the laws of physics. Compared to traditional neural networks \cite{reveiw1, reveiw2}, PINNs incorporate physics governing equations into the loss function, enabling them to learn a more generalized model with fewer labeled data. Neural networks were first used to solve partial differential equations in the 1990s \cite{old-phy}. However, this approach did not gain much attention due to the limitations of hardware and computing methods. With the development of deep learning, Raissi et al. proposed the framework of PINNs, leading to significant subsequent research in this emerging cross-disciplinary field \cite{PINN} and becoming a hotspot in scientific computation and artificial intelligence.

There are several studies have shown that neural networks can converge to PDEs' solution under certain conditions. For instance, Shin et al. analyzed the consistency of using PINN to solve PDEs. They demonstrated that the upper bound on the generalization error is controlled by the training error and the number of training data in Holder continuity assumptions \cite{shin}. Similarly, Mishra et al. established a theory of an abstract framework that uses PINN to solve forward PDEs problem \cite{mishra}. They obtained a similar error estimation result under a weaker assumption and generalized it to the inverse problem. Despite these promising results, the accuracy of PINNs is often not satisfactory enough, and they are difficult to train or fail to converge in certain situations \cite{theory1}.

To address the challenges of PINNs, several researchers have proposed new models and training algorithms to refine the original method \cite{ijcai1, ijcai2, cpinn, xpinn, gpinn, adp, bpinn}. For instance, Jagtap et al. proposed a nonlinear conservation law physics-informed neural network in the discrete domain (c-PINN) \cite{cpinn}, which divides the solution domain into sub-domains connected by a conservation law. This strategy can improve efficiency and accuracy. However, the connection of the sub-domain needs to abide by conservation law. Therefore, Ameya et al. proposed the X-PINN model, which further expands the c-PINN to overcome the limitation of conservation law \cite{xpinn}. In another approach, Jeremy et al. proposed the g-PINN, which includes higher derivatives of equations in the loss function to provide extra information to train \cite{gpinn}. Their experiments suggest that these derivative forms can help the model converge. To account for different learning difficulties in different regions of the training domain, Wu et al. proposed a self-adapted points sampling algorithm that increases the sampling rate in high physics loss regions \cite{adp}. Since physics data in many fields often have some noise, Yang et al. proposed a Bayes physics-informed neural network (B-PINN) to predict accuracy values in noisy data \cite{bpinn}. Furthermore, PINNs have successfully been applied to practical problems in various fields, including fluid mechanics, mechanics of materials, power systems, and biomedical \cite{app1, app2, app3}. However, few studies have investigated applying semi-supervised learning methods to PINNs.

Self-training, also known as self-labeling or pseudo-labeling training, is a primary method in semi-supervised learning \cite{semi1, semi2, semi3, semi4, semi5}. The main idea behind self-training is to use the network's prediction as pseudo labels and convert unsupervised data into supervised data. This approach allows the model to learn from unlabeled data, which suits the scenarios where label generation is costly, such as solving PDEs. In this paper, we employ the self-training mechanism to improve the accuracy of the PINN network by enhancing its physics learning process.

\begin{figure*}[htbp]
\centering
\includegraphics[width=0.75\linewidth]{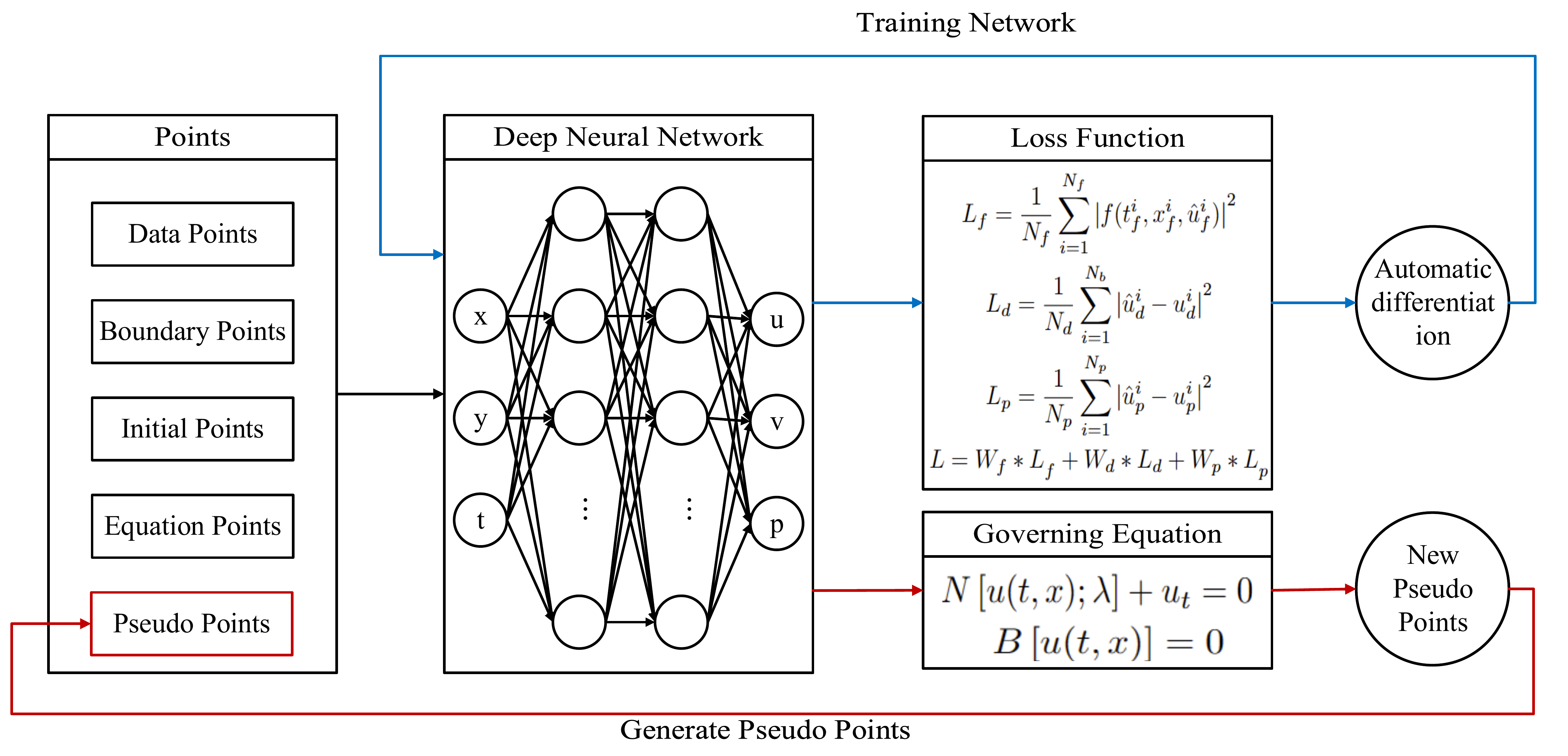}
\caption{The architecture of ST-PINN. The blue line displays the training process, while the red line presents the pseudo label generating process.} \label{architecture}
\end{figure*}

\section{Method}

We first define the general form of PDEs problem with boundary conditions as following:
\begin{gather}
     N\left[u(t,x);\lambda\right]+u_t=0,x\in\Omega\subset\mathbb{R}^d,t\in[0,T]\label{eq1} \\
     B\left[u(t,x)\right]=0,\partial\Omega\label{eq2}
\end{gather}
where $u(t,x)$ denotes the solution to the PDE at time $t$ and location $x$; $\lambda$ represents the unknown parameters in the equation; $u_t$ is the equation forcing function; $N[\cdot]$ is a nonlinear differential operator; and $B(\cdot)$ represents the boundary conditions, which can be Dirichlet, Neumann, or mixed. In PINNs, the initial condition can be treated as a Dirichlet boundary condition.

\subsection{The Overview of the Framework}

Fig.~\ref{architecture} shows the architecture of the proposed network, which uses time $t$ and spatial coordinates (e.g. $x$, $y$) as inputs and predicts the physical fields (e.g. $u$ and $v$ for velocity fields in two directions and $p$ for the pressure field) as outputs. There are two major processes involved in one iteration: training the network and generating pseudo labels. The blue line represents the training process of the network model, which is similar to the original PINNs. The sample points and boundary points are input to the network to generate the prediction, which computes the physics residual loss of the governing equations. Although PINNs can learn without supervised data, they usually require a few labeled intra-domain points to speed up convergence and improve accuracy. The unsupervised data learn the physical information by combining governing equations into the loss function and using the back-propagation algorithm to train the network. Some articles refer to combining governing equations into the loss function as a ``soft constrain'' because there is no guarantee that the network output can satisfy the PDEs. In our paper, we train the network by directly learning the residual between the pseudo labels and the prediction, which can learn more physical information.

To further improve the training process and accuracy, we introduce a novel pseudo-labeling technique that integrates the generated pseudo labels into the training process. As shown in Fig.~\ref{architecture}, the red line illustrates the main difference between the original PINN and ST-PINN. In ST-PINN, after several iterations, the network predicts all the sample points and uses these predictions to compute the residual loss of equation. By sorting the residual loss, the network selects part sample points with small equation residual loss for pseudo labeling, which will be trained as supervised data in the next iteration. At the end of each iteration, the proposed method generates new pseudo points through the above process. Therefore, the number of pseudo points will gradually increase during training. If the point-generating strategy is suitable, there will be a positive feedback loop between the pseudo points generation and the network training. These pseudo points can improve the accuracy of the network, and the more accurate the network becomes, the more pseudo points it can generate, leading to convergence.

However, theoretical analysis of the self-training and the pseudo-label points requires extensive work, which is beyond the scope of this study.

\subsection{Loss Function}
The training points of the network can be divided into several parts, each with its corresponding loss function. The first part consists of sample points randomly sampled from the spatiotemporal domain to learn the physics information.  The residual loss of governing equations shown in \eqref{eq3}, where $f(t,x,u)=N\left[u(t,x);\lambda\right]+u_t$ is the governing equations, $t_f$, $x_f$ represent the randomly sampled points on the training domain, and ${\hat{u}}_f$ is the prediction of the neural network. The network learns the physical information by minimizing the mean squared error of the equation residual, which is a crucial part of traditional PINNs. We aim to make the output of PINNs as compliant with the equation as possible by minimizing $f$.
\begin{gather}
     L_f=\frac{1}{N_f}\sum_{i=1}^{N_f}{|f(t_f^i,x_f^i,\hat{u}_f^i)|}^2 \label{eq3} \\
     L_d=\frac{1}{N_d}\sum_{i=1}^{N_b}{|{\hat{u}}_d^i-u_d^i|}^2 \label{eq4} \\
     L_p=\frac{1}{N_p}\sum_{i=1}^{N_p}{|{\hat{u}}_p^i-u_p^i|}^2 \label{eq5} \\
     L={W_f\ast L}_f+W_d\ast L_d+{W_p\ast L}_p \label{eq6} 
\end{gather}

The second part comprises labeled data, which includes boundary and initial points, and sometimes a small number of intra-domain points. These points can be generalized as data points and calculated using \eqref{eq4}, where ${\hat{u}}_d$ and $u_d$ represent the predictions and corresponding labels, respectively. The initial and boundary conditions are essential for PINNs to find the correct solution. Although PINNs can learn the PDE solution without supervised data, adding a few labeled intra-domain points can improve convergence speed and accuracy.

The third part, the loss function for pseudo points defined in \eqref{eq5}, is the main difference between the original PINNs. Here, ${\hat{u}}_p$ and $u_p$ denote the predictions and corresponding pseudo labels, respectively. These pseudo labels are generated in previous iterations and represent high-confidence predictions. It is worth noting that the representation of the pseudo points and data points is similar, and they can use the same loss function defined in Equation \eqref{eq4}. However, it is beneficial to separate them into distinct losses in practice and adjust the loss weights to improve accuracy. Furthermore, splitting these loss functions helps to analyze the behavior of the self-training mechanism. Finally, we compute the total loss by taking a weighted sum of the above losses, as shown in Equation \eqref{eq6}.

\subsection{The Pseudo Points Generation Strategy}

Selecting unsupervised data to generate pseudo labels is a crucial and subjective procedure in self-training \cite{semi2}. In classification problems, the traditional approach for pseudo-labeling is to set a threshold on the classification confidence. At the end of each iteration, the unlabeled data with high classification confidence is assigned pseudo labels and included in the training set. As the training progresses, the number of pseudo points increases, and the network output becomes more accurate. Eventually, the unsupervised data becomes labeled, effectively transforming it into supervised data.

However, solving PDEs is not a classification problem, and self-training cannot be applied to PINNs directly. Firstly, using the threshold is not always convenient since the confidence depends on the physics equations. Different problems need different threshold values, and even the same PDE with specific initial or boundary conditions needs to change thresholds. Moreover, sometimes it is challenging to choose the thresholds without experiments. Therefore, the ST-PINN uses a qualitative indicator to select the pseudo points, such as choosing the top $q$\% of sample points as the pseudo points. 

The second problem is that generating pseudo points every iteration is unnecessary, leading to useless overhead and decreased efficiency. To address this, we propose reducing the frequency of pseudo point updates in ST-PINN, with the update frequency controlled by a hyperparameter $p$. This approach strikes a balance between efficiency and accuracy. During training, the network parameters are updated at every iteration, and the pseudo points generated by the previous network may change in the new network. To ensure accurate propagation of physical information and avoid error accumulation, ST-PINN updates or replaces all pseudo points with the output of the new network, which differs from traditional self-training methods.

Another challenge with selecting pseudo points based solely on the PDE is that adding the pseudo points residual into the loss function can cause fluctuations in the training process. To improve stability, ST-PINN introduces a stable coefficient, denoted by $r$, which measures the number of consecutive times a candidate point remains stable before attaching a pseudo label. Specifically, during training, we maintain a window to keep track of the flags of all predictions, which records the number of consecutive candidate times. Pseudo labels are only attached once the flags of corresponding points surpass $r$. By setting an appropriate value for $r$, the network can improve the quality of pseudo labels and ensure the stability of the training process. The stable coefficient also serves as a mechanism to control the number of pseudo points: at the beginning of training, it limits the number of pseudo points to warm up the network, and as the iteration count increases, it allows the network to generate more and more pseudo points, acting as a pseudo points schedule.

In general, our pseudo-labeling strategy introduces three hyperparameters: the update frequency $p$, the maximum rate $q$, and the stable coefficient $r$. The pseudo points selection algorithm is described in Algorithm 1, which completes one iteration in five steps. First, the network generates predictions from sample points and computes their residual loss of equation (lines 3-4). Then, the ST-PINN sorts the candidate points by the loss and selects the top $q$\% sample points as pseudo point candidates (line 6). After that, the proposed model updates the consecutive time flags and chooses candidate points if the flag surpasses the stable coefficient $r$ as pseudo points (lines 7-14). Once the pseudo point generation completes, the ST-PINN resets the flag of non-candidates (lines 15-17). In Section 4, we provide a detailed evaluation of our method.

\begin{algorithm}[h]
  \caption{Generation of pseudo points}
  \begin{algorithmic}[1]
    \State ${t_p,x_p,u}_p\gets\emptyset$
    \For{$i=0$ to $N_f-1$}
      \State ${{\hat{u}}_f^i\gets h}^{\left(k\right)}\left(t_f^i,x_f^i\right)$
      \State $L_f^i\gets f\left(t_f^i,x_f^i,{\hat{u}}_f^i\right)$ 
    \EndFor
    \State idx = argPartition($L_f,t_f,x_f,{\hat{u}}_f,q$)
    \For{$i=0$ to $N_f\times q-1$}
      \State flag[idx[i]] $+= 1$
        \If{flag[idx[i]] $\textgreater\ r$}
            \State $t_p\gets t_p\cup t_f^i$
            \State $x_p\gets x_p\cup x_f^i$
            \State $u_p\gets u_p\cup{\hat{u}}_f^i$
        \EndIf
    \EndFor
    \For{$i=N_f\ast q$ to $N_f-1$}
      \State flag[idx[i]] = 0
    \EndFor
    \label{algroithm}
  \end{algorithmic}
\end{algorithm}

\section{Experimental Results}

We conduct a series of experiments on five different PDEs from different fields and scenarios, including the Burgers equation, Different-Reaction equation, Different-Sorption equation, Shallow-Water equation, and Compressible Navier-Stokes equation. All the training data are downloaded from the PDEBench dataset \cite{pdebench}. The setting of equations and parameters are default. Our deep learning framework is TensorFlow 1.15. The accelerator is NVIDIA P100.

\begin{figure*}[htbp]
\centering
\includegraphics[width=\linewidth]{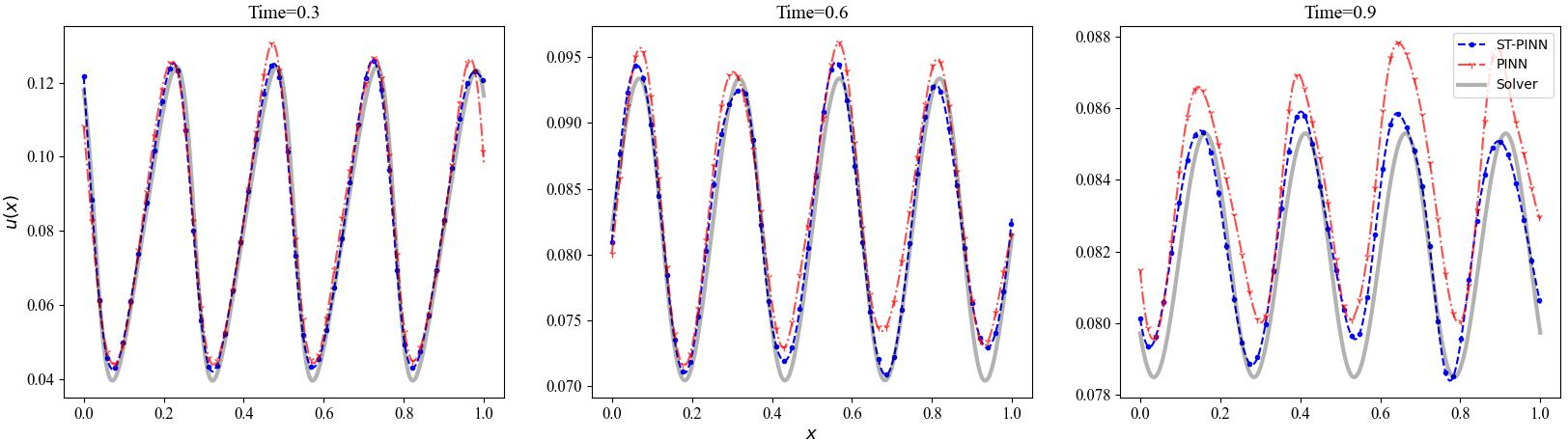}
\caption{The prediction of PINN and ST-PINN at three different times (t=0.3, t=0.6, and t=1).} \label{Burgers1D}
\end{figure*}

\subsection{Burgers Equation}

The Burgers equation and the corresponding initial conditions can be described by (7) and (8), which model the non-linear behavior and diffusion process in fluid dynamics:
\begin{equation}
\partial_tu\left(t,x\right)+\partial_x\left(u^2(t,x\right)/2)=\nu/\pi\partial_{xx}u\left(t,x\right),
\label{eq7}
\end{equation}
\begin{equation}
    u\left(0,x\right)=u_0\left(x\right),\ \ x\in(0,1) \label{eq8}
\end{equation}
where $\nu=0.01$ represents the diffusion coefficient. The boundary condition is periodic. \eqref{eq9} describes the initial condition, which is a superposition of sinusoidal waves:
\begin{equation}
    u_0\left(x\right)=\sum_{k_i=k_1,...,k_N}{A_i\sin{{(2\pi{n_i}/L_x)x+\emptyset_i}}} \label{eq9}
\end{equation}
where $L$ is the calculation domain size; $n_i$, $A_i$ and $\emptyset_i$ are random sample values. $n_i$ is a random integer in $[1,8]$; $A_i$ is a random float number uniformly chosen in $[0,1]$, and $\emptyset_i$ is a randomly chosen phase in $(0,2\pi)$. Note that $N=2$ in this equation.

The networks were trained over a spatiotemporal domain of $[0,1]\times[0,2]$ and discretized into $N_x\times N_t=1024\times256$ points. Both PINN and ST-PINN models utilized fully connected neural networks with four layers and 32 cells in each layer. We select the network structure by grid search. The number of boundary points was 512, and the number of initial points was 1024. We also put 1000 intra-domain labeled data points to improve training efficiency and accuracy. Both networks were trained for 20,000 iterations using the Adam optimizer, with a learning rate of $10^{-3}$ and the activation function set to $tanh$.

Because the total number of points in the training domain ($1024\times256=262,144$) was too large to input into the network at once, we randomly sampled 20,000 points for each iteration, which can be considered a mini-batch approach with a batch size of 20,000. Only sample points and pseudo points were generated within each batch, with other points (boundary, initial, and data points) included in training for every iteration.

It's worth noting that unless specifically mentioned, the training environment and network configuration were kept the same for both PINN and ST-PINN models, ensuring a fair comparison of their solution prediction accuracy.

As for the pseudo points generation strategy, the network updates at most 20\% of the sample points every 100 iterations, and the stable coefficient is set as 10, which is a relatively conservative setting. Therefore, the number of pseudo points is small at the beginning of the training. This setting ensures the network does not deviate from the solution even if they learn the wrong information.

Fig.~\ref{Burgers1D} illustrates the prediction of the PINN and ST-PINN models at three different times (t=0.3, t=0.6, and t=1), where the blue line represents the ST-PINN and the red line expresses the PINN. Both networks can fit the solution but have difficulties predicting the peaks and troughs, particularly the PINN. This phenomenon becomes more evident as time increases, and at t=0.9, the PINN can only forecast the form of the prediction without fitting the wave accuracy. In contrast, the result obtained with ST-PINN is much better, with its prediction being nearer to the solution in almost every peak and trough. The relative L2 error for ST-PINN (6.60e-2) is also better than that for PINN (8.01e-2). One possible reason for this phenomenon is that wave peaks and troughs are complex parts of the solution to PDEs and are challenging to learn. Therefore, the self-training mechanism in ST-PINN helps it to learn more physical information.

\subsection{Diffusion-Reaction Equation}

The one-dimensional diffusion-reaction equation and its corresponding initial conditions are expressed as follows: 
\begin{equation}
    \partial_tu\left(t,x\right)-\upsilon\partial_{xx}u\left(t,x\right)-\rho u\left(1-u\right)=0,
\label{eq10}
\end{equation}
\begin{equation}
u(0,x)=u_0(x),\ \ \ x\in(0,1) \label{eq11}
\end{equation}
where $\upsilon$ is the diffusion coefficient, and $\rho$ denotes mass density, which is set as 0.5 and 1, respectively. The boundary conditions and initial conditions are the same as those for Burgers equation, with periodic boundary conditions and the superposition of sinusoidal waves defined in \eqref{eq9} used as the initial condition. This equation combines a diffusion process and a rapid evolution from a source term, making it a challenging problem that can measure the network's ability to capture very swift dynamics. 

The spatiotemporal domain for training is $[0,1]\times[0,1]$, and is discretized into $N_x\times N_t=1024\times256$ points. The networks are trained using the Adam optimizer for 20,000 iterations, followed by further refinement using the Limited-memory Broyden-Fletcher-Goldfarb-Shanno Bound (L-BFGS-B) optimizer for a maximum of 5,000 iterations. L-BFGS-B is a second-order optimizer that can effectively decrease loss with limited memory expenditure. Other settings (e.g. the network structure and pseudo point generation strategy) are similar to the Burgers equation.

\begin{figure}[htbp]
\centering
\includegraphics[width=0.85\linewidth]{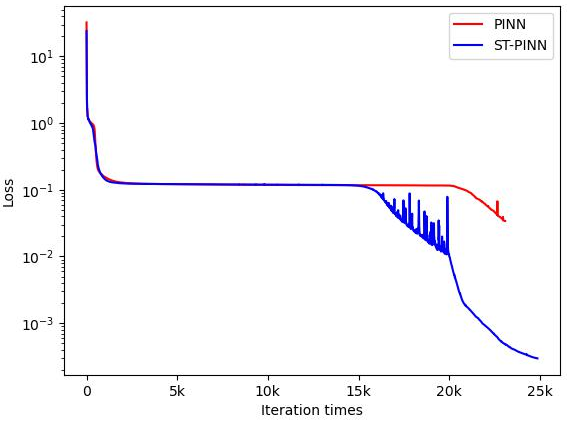}
\caption{The variation of the loss function.} \label{Diff-React-1D}
\end{figure}

Fig.~\ref{Diff-React-1D} displays the variation of the loss function during the training process for the diffusion-reaction equation. The red line represents the loss of PINN, while the blue line corresponds to ST-PINN. At the beginning of training, the loss variation between PINN and ST-PINN is minimal. However, after 15k iterations, the self-training mechanism generates enough pseudo points to influence the training process, which helps the network learn more physical information, leading to a faster reduction of the loss to a lower level in ST-PINN. When using the L-BFGS-B optimizer with a total of 20k iterations, the loss in ST-PINN descends even faster. This phenomenon highlights the advantage of self-training in improving performance.

From Fig.~\ref{Diff-React-1D}, we can see that the loss in PINN converges early, reaching its minimum value at approximately 23k iterations. In contrast, the loss in ST-PINN continues to decrease until the training stops. As a result, the final L2 error in ST-PINN (9.61e-3) is better than that in PINN (4.22e-2).

\subsection{Different-Sorption Equation}

The one-dimensional diffusion-sorption equation models a diffusion process that is retarded by a sorption process:
\begin{equation}
    \partial_tu\left(t,x\right)\!=\!D/R\left(u\right)\partial_{xx}u\left(t,x\right)\!,x\!\in\!(0,1),t\!\in\!(0,500]  \label{eq12}
\end{equation}
where $D=5\times{10}^{-4}$ is the effective diffusion coefficient and $R(u)$ describes the retardation factor representing the sorption that hinders the diffusion process:
\begin{equation}
 R(u)=1+\frac{1-\emptyset}{\emptyset}\rho_skn_fu^{n_f-1} \label{eq13}
\end{equation}
Here, $\emptyset=0.29$ represents the porosity of the porous medium; $\rho_s=2888$ denotes the bulk density; $k=3.5\times{10}^{-4}$ is Freundlich’s parameter; and $n_f=0.875$ is Freundlich’s exponent. The boundary conditions are defined by:
\begin{gather}
u(t,0)=1.0 \label{eq14} \\
u(t,1)=D\partial_xu\left(t,1\right) \label{eq15}
\end{gather}
The initial condition is generated by random values with a uniform distribution. The discrete training domain is $N_x\times N_t=1024\times101$.

\begin{figure}[htbp]
\centering
\includegraphics[width=0.75\linewidth]{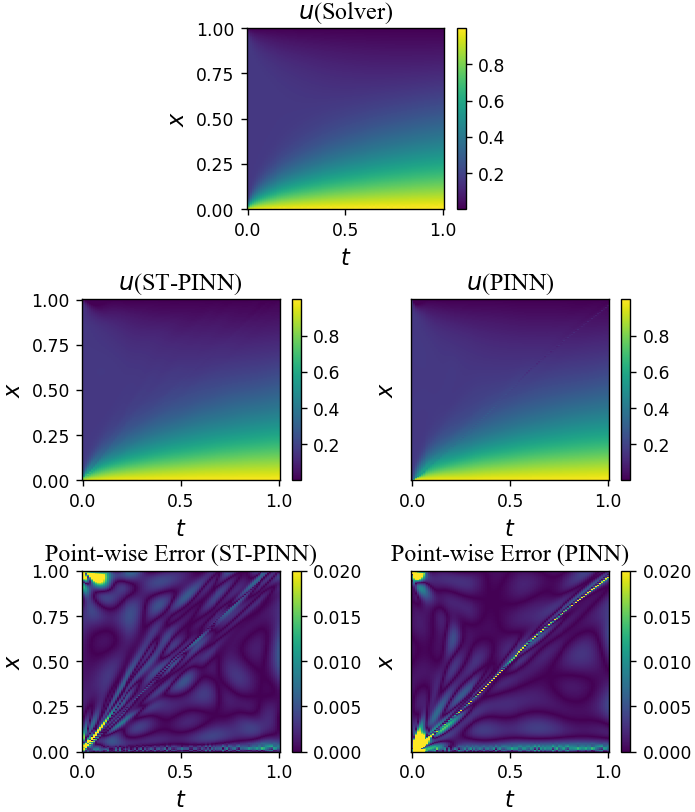}
\caption{The reference solution, prediction, and the point-wise error of PINN and ST-PINN.} \label{Diff-Sorb-1D}
\end{figure}

For neural networks, we used the same architectures for both PINN and ST-PINN as we did for the Burgers equation and Diffusion-Reactive equation. Both models were trained for 20,000 iterations using the Adam optimizer. The relative L2 error in ST-PINN (9.63e-3) is better than PINN (2.45e-2). Fig.~\ref{Diff-Sorb-1D} shows the reference solution, prediction, and point-wise error in PINN and ST-PINN. The first two rows show the reference solution and prediction in both networks, which are similar to each other, indicating that they can predict the physics fields accurately. The third row shows the point-wise error compared to the reference solution in both networks. In PINN, the point-wise error at the bottom ($x=0$) and diagonal ($x=0.5t$) shows a significant deviation. However, this phenomenon is less pronounced in ST-PINN. Thus, the ability of ST-PINN to capture the physics details is better.

\subsection{Shallow-Water Equations}

The two-dimensional shallow-water equation is expressed as \eqref{eq16} - \eqref{eq18}:
\begin{gather}
    \partial_th+\partial_xhu+\partial_yhu=0 \label{eq16} \\
    \partial_thu+\partial_x\left(u^2h+\frac{1}{2}g_rh^2\right)=-g_rh\partial_xb \label{eq17} \\
    \partial_thv+\partial_y\left(v^2h+\frac{1}{2}g_rh^2\right)=-g_rh\partial_yb \label{eq18} 
\end{gather}
where $u$ and $v$ denote velocities in the horizontal and vertical directions, $h$ describes the water depth (the main prediction in this problem), and $g_r=1.0$ is the gravitational acceleration. These equations are derived from the general Navier-Stokes (N-S) equations and are widely used in modeling free-surface flow problems.

The dataset presented in PDEBench is a 2D radial dam break scenario, which describes the evolution process of a circular bump with initialized water height in the center of a square domain. The initial condition for $h$ is given by:
\begin{equation}
    h=\left\{\begin{matrix}2.0,&for\ r\ <\sqrt{x^2+y^2}\\1.0,&for\ r\ \geq\sqrt{x^2+y^2}\\\end{matrix}\right. \label{eq19}
\end{equation}
where $r=0.1$. The training spatial dimension is $\Omega=[-2.5,2.5]\times [-2.5,2.5]$, while the temporal dimension is $T=[0,1]$. The dataset is discretized into $N_x \times N_y \times N_t = 128 \times 128 \times 101$.

Unlike the one-dimensional case study introduced above, the two-dimensional problem is more complex. Therefore, we use a deeper network with more neural cells for solving these equations. The number of boundary points and initial points is also increased, with 1000 boundary points, 5000 initial points, and an additional 1000 intra-domain labeled data points used to train both networks. Both networks are trained using the Adam optimizer with 30k iterations. The learning rate for the first 10k iterations is set to 5e-3, the learning rate for the second 10k iterations is set to 1e-3, and the learning rate for the last 10k iterations is set to 5e-4. 

\begin{figure}[htbp]
\centering
\includegraphics[width=0.74\linewidth]{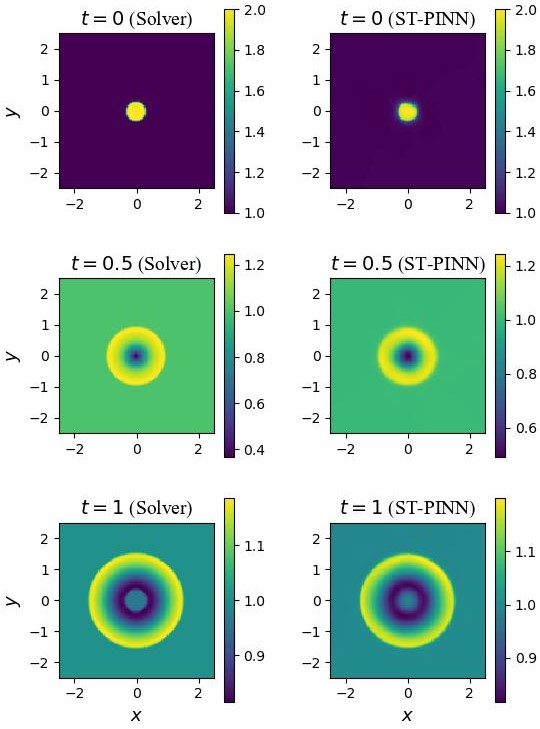}
\caption{The reference solution and prediction of ST-PINN.} \label{SWD2D}
\end{figure}

One significant difference in this case study is that the dataset only provides labeled data for $h$, while the labels for $u$ and $v$ are unavailable. However, the shallow-water equation contains physical information about $u$ and $v$. Therefore, the ST-PINN can predict them and include the no-reference points in the self-training process. Fig.~\ref{SWD2D} displays the prediction of the ST-PINN and the corresponding reference. By incorporating these unlabeled physics data, ST-PINN can accurately forecast the 2D physics fields. The L2 error of $h$ in ST-PINN (1.14e-2) is better than that of PINN (1.50e-2).

\subsection{Compressible Navier-Stokes Equations}
The two-dimension compressible Navier-Stokes (N-S) equations is expressed as \eqref{eq20} - \eqref{eq22}:
\begin{gather}
    \partial_t\rho+\nabla\cdot(\rho\mathbf{v})=0 \label{eq20} \\
    \rho(\partial_t\mathbf{v}+\mathbf{v}\!\cdot\!\nabla\mathbf{v})=-\nabla p+\eta\Delta\mathbf{v}+(\zeta+\eta/3)\nabla(\nabla\cdot\mathbf{v}) \label{eq21} \\
    \partial_t\left[\epsilon+\frac{\rho v^2}{2}\right]+\nabla\cdot\left[\left(\epsilon+p+\frac{\rho v^2}{2}\right)\mathbf{v}-\mathbf{v}\cdot\sigma^\prime\right]=0 \label{eq22} 
\end{gather}
In the compressible Navier-Stokes (N-S) equations, $\rho$ represents the mass density, $v$ denotes the velocity, and $p$ expresses the gas pressure. The internal energy $\epsilon$ is given by $\epsilon=p/(\Gamma-1)$, where $\Gamma=5/3$ is the specific heat ratio. The Mach number is set to 0.1, and the viscous stress tensor $\sigma^\prime$ accounts for the shear and bulk viscosity, with values of $\eta=0.1$ and $\zeta=0.1$, respectively. These equations are subject to periodic boundary conditions and random initial conditions. The N-S equation is a complicated equation that governs the mechanical law of viscous fluid flow,  making it a rigorous test for the network's ability to understand and represent complex physical information.

The dataset used in this problem is discrete into $N_x\times N_y\times N_t=128\times 128\times 21$, with a training spatial dimension of $\Omega=[-2.5,2.5]\times[-2.5,2.5]$ and a temporal dimension of $T=[0,1]$. The input to the network is the coordinates $(x,y,t)$, and the output is the physical fields $(\rho,u,v,p)$ at those coordinates. The network architecture consists of eight layers with 64 cells in each layer, similar to the network used for the shallow-water equation. Note that the scales of $\rho$ and $p$ are greater than those of $u$ and $v$. We use mean square error (MSE) to evaluate $u$ and $v$ and the L2 error to evaluate $\rho$ and $p$. Tab.~\ref{ns-result} shows the final results. Overall, the performance of ST-PINN is better than PINN in terms of $\rho$, $u$, and $v$, demonstrating the network's ability to handle complex problems.

\begin{table}[htbp]
\caption{The MSE and related L2 error of ST-PINN and PINN}
\begin{center}
\begin{tabular}{cccccc}
\toprule
    & & $\rho$              & $u$               & $v$              & $p$              \\
\midrule
\multirow{2}{*}{MSE} & PINN    & 4.50e-2               & 6.76e-3               & 6.56e-3               & 3.16e-2       \\
                     & ST-PINN & 2.67e-2               & 5.11e-3               & 5.21e-3               & 3.25e-2       \\
\midrule
\multirow{2}{*}{L2}  & PINN    & 4.73e-3               & 1.05e-0               & 1.16e-0               & 1.28e-2       \\
                     & ST-PINN & 2.81e-3               & 7.94e-1               & 9.16e-1               & 1.31e-2       \\                     \bottomrule
\end{tabular}
\label{ns-result}
\end{center}
\end{table}

\section{CONCLUSION}
We propose ST-PINN, which incorporates the self-training mechanism into physics-informed learning. The core of ST-PINN is to embed the residual of pseudo points into the loss function, thus extending the physical information by directly learning from the residual. The ST-PINN selects the most confident sample points as pseudo points using the physics governing equation and has three hyperparameters that control the number of pseudo points and the frequency of pseudo-label generation. These hyperparameters stabilize the training process and balance accuracy and efficiency. To our best knowledge, this is the first research to apply the semi-supervised learning method to physics-informed learning. We conducted experiments on five PDEs in various scenarios and applications. Compared to the original PINN, the proposed network improved accuracy by about 1.33$\times$-2.54$\times$. These results demonstrate that the self-training mechanism can benefit network convergence and improve prediction accuracy. In the future, we plan to optimize the network architecture to enhance performance and generalization ability. Investigating relevant mathematical demonstrations of ST-PINN is also a future focus of our research. 


\bibliographystyle{STPINN}
\bibliography{STPINN}

\end{document}